%% file: acl2023.tex
\pdfoutput=1

\documentclass[11pt]{article}

\usepackage{ACL2023}

\usepackage{times}
\usepackage{latexsym}

\usepackage[T1]{fontenc}

\usepackage[utf8]{inputenc}

\usepackage{microtype}

\definecolor{verde}{rgb}{0.0, 0.5, 0.0}

\usepackage{todonotes}
\usepackage{amsmath} 
\usepackage[shortlabels]{enumitem}
\usepackage{comment}
\usepackage{multirow}
\usepackage{booktabs}
\usepackage{float}

\title{{O}n the {I}nterpretability and {S}ignificance of {B}ias {M}etrics in {T}exts: a {PMI}-based {A}pproach}

\author{
    Francisco Valentini$^{1,2}$ \,
    Germán Rosati$^{3}$ \,
    Damián Blasi$^{4}$ \,
    \textbf{Diego Fernandez Slezak}$^{1,5}$  \\
    \textbf{Edgar Altszyler}$^{1,2}$ \\ \\[-.1cm]
    $^{1}$Instituto de Investigación en Ciencias de la Computación, CONICET-UBA, Argentina \\
    $^{2}$Maestría en Data Mining, Universidad de Buenos Aires (UBA), Argentina \\
    $^{3}$CONICET. Escuela IDAES, Universidad Nacional de San Martín, Argentina \\ 
    $^{4}$Harvard University, USA \\
    $^{5}$Departamento de Computación, FCEyN, UBA, Argentina \\
    \small{\texttt{fvalentini@dc.uba.ar}},\, 
    \small{\texttt{grosati@unsam.edu.ar}},\, 
    \small{\texttt{dblasi@fas.harvard.edu}},\, \\[-.1cm]
    \small{\texttt{dfslezak@dc.uba.ar}},\, 
    \small{\texttt{ealtszyler@dc.uba.ar}}\\
}

\begin{document}
\maketitle
\begin{abstract}
In recent years, word embeddings have been widely used to measure biases in texts. Even if they have proven to be effective in detecting a wide variety of biases, metrics based on word embeddings lack transparency and interpretability. We analyze an alternative PMI-based metric to quantify biases in texts. It can be expressed as a function of conditional probabilities, which provides a simple interpretation in terms of word co-occurrences. We also prove that it can be approximated by an odds ratio, which allows estimating confidence intervals and statistical significance of textual biases. This approach produces similar results to metrics based on word embeddings when capturing gender gaps of the real world embedded in large corpora.\footnote{Code for the paper is available at \url{https://github.com/ftvalentini/BiasPMI}}
\end{abstract}

\section{Introduction}

Word embedding-based approaches have been used for detecting and quantifying gender, ethnic, racial, and other stereotypes present in corpora. While some research has focused on investigating biases by training embeddings on a specific corpus of interest \citep{garg2018word, kozlowski2019geometry, lewis2020gender, charlesworth2021gender}, others have employed pretrained word embeddings to assess potential biases inherent in the training corpus \citep{caliskan2017semantics, garg2018word, defranza2020language, jones2020stereotypical}.


Though not as popular, Pointwise Mutual Information (PMI) is a measure of word similarity which has also been used to study biases \citep{galvez2018half,bordia2019identifying,aka2021measuring}. However, the statistical properties and advantages of this measure as compared to the widely used word embeddings have not been studied yet.

In this article we study a PMI-based metric to measure bias in corpora and explain its statistical and interpretability benefits, which have been overlooked until now. Our contributions are as follows: (1) We show the PMI-based bias metric can be approximated by an odds ratio, which makes computationally inexpensive and meaningful statistical inference possible. (2) We provide evidence that methods based on GloVe, skip-gram with negative sampling (SGNS) and PMI produce comparable results when the biases measured in large corpora are compared to empirical information about the world. (3) We contend that the PMI-based bias metric is substantially more transparent and interpretable than the embedding-based metrics.

\begin{description}[wide,itemindent=\labelsep]

\item[Scope:] The detection and mitigation of bias in models is a research topic that is beyond the scope of this paper. Our paper’s contribution focuses on the measurement of bias in raw corpora (not models), which is a relevant task in Computational Social Science. 

\end{description}

\section{Background}

Consider two sets of context words $A$ and $B$, and a set of target words $C$. Textual bias measures quantify how much more the words of $C$ are associated with the words of $A$ than with those of $B$. Most metrics can be expressed as a difference between the similarities between $A$ and $C$, on the one hand, and $B$ and $C$, on the other: 
\begin{equation} \label{eq:bias_measure_generico}
    \text{Bias} = \text{sim}(A,C) - \text{sim}(B,C)
\end{equation}
For instance, to estimate the female vs. male gender bias of occupations, context words are often gendered pronouns or nouns, e.g., \emph{A = \{she, her, woman,..\}} and \emph{B = \{he, him, man,...\}}; whereas $C$ is usually considered one word at a time, estimating for each specific job (\emph{nurse}, \emph{doctor}, \emph{engineer}, etc.) the relative association to $A$ and $B$. 

One particularly popular metric which uses word embeddings (WE) is that of \citet{caliskan2017semantics}:
 \begin{equation}\label{eq:BiasCaliskan} 
\text{Bias}_{\text{WE}} = 
\frac{ \underset{a \in A}{\mathrm{mean}} \; \text{cos}(v_a,v_c) - 
\underset{b \in B}{\mathrm{mean}} \; \text{cos}(v_b,v_c) }
{ \underset{x \in A\cup B} {\operatorname{std\_dev}} \; \text{cos}(v_x,v_c) }
\end{equation}
where $v_i$ stands for the word embedding of word $i$ and $\text{cos}(v_i,v_j)$ is the cosine similarity between vectors.

Permutations tests that shuffle context words have been used to calculate the statistical significance of $\text{Bias}_{\text{WE}}$ \citep{caliskan2017semantics,charlesworth2021gender}. These tests permute the words from $A$ and $B$ repeatedly and compute the bias metric in each iteration to simulate a null distribution of bias. The two-tailed p-value is calculated as the fraction of times the absolute value of bias from the null distribution is equal to or greater than the one observed \citep{north2002note}.

With a similar re-sampling approach, bootstrap can also be performed \citep{garg2018word}. The bootstrap distribution is obtained by calculating the bias metric over many bootstrap samples from $A$ and $B$, sampled separately for each group. The standard error of bias is then estimated as the sample standard deviation of the bootstrap distribution, and the quantiles of the distribution are used to obtain percentile confidence intervals \citep{davison1997bootstrap}.

\section{Bias measurement with PMI}

Here we introduce a bias metric that follows equation \ref{eq:bias_measure_generico} but uses Pointwise Mutual Information (PMI) \cite{church1990word} as a measure of word similarity: 
\begin{equation}\label{eq:BiasPMI}
 \text{Bias}_{\operatorname{PMI}} =  
                \operatorname{PMI}(A,C) - \operatorname{PMI}(B,C)
\end{equation}

PMI measures the first-order association between two lists of words $X$ and $Y$:
\begin{equation}\label{eq:PMI_lists}
    \operatorname{PMI}(X,Y) 
        = \log \, \frac{P(X,Y)}{P(X)P(Y)}
        = \log \, \frac{P(Y|X)}{P(Y)},
\end{equation}
where $P(X,Y)$ is the probability of co-occurrence between any word in $X$ with any one in $Y$ in a window of words, and $P(X)$ and $P(Y)$ are the probability of occurrence of any word in $X$ and any word in $Y$, respectively. Equation \ref{eq:PMI_lists} shows PMI can be expressed as the ratio between the probability of words in $Y$ co-occurring with words in $X$, and the probability of words in $Y$ appearing in any context.

\subsection{Approximation of the PMI-based bias by log odds ratio}\label{sec:odds_ratio}

Combining equations \ref{eq:BiasPMI} and \ref{eq:PMI_lists}, the PMI-based bias can be written as a ratio of conditional probabilities, which can be estimated via maximum likelihood using the co-occurrence counts from the corpus:
\begin{equation}\label{eq:PMI_estimate}
 \text{Bias}_{\operatorname{PMI}} 
    = \log \, \frac{P(C|A)}{P(C|B)}
    = \log \, \frac{\frac{f_{A,C}}{f_{A,C}+f_{A,nC}}}{\frac{f_{B,C}}{f_{B,C}+f_{B,nC}}},
\end{equation}
where $f_{A,C}$ and $f_{B,C}$ represent the number of times words in $C$ appear in the context of words in $A$ and $B$, respectively, and $f_{A,nC}$ and $f_{B,nC}$ represent how many times words not in $C$ appear in the context of $A$ and $B$, respectively. See contingency table in Appendix \ref{sec:contingency_table} for reference.

$\text{Bias}_{\operatorname{PMI}}$ is not computable if $f_{A,C}=0$ or $f_{B,C}=0$. We address this by adding a small value $\epsilon$ to all co-occurrences in the corpus \citep{jurafsky2000speech}. 

For most practical applications, co-occurrences between words not in a group (most of the vocabulary) and a group of specific words are larger than the co-occurrences between two groups of specific words. More precisely:
\begin{equation}\label{eq:approx}
    {f_{B,nC}\gg f_{B,C}, \: f_{A,nC}\gg f_{A,C}}.
\end{equation}
Thus:
\begin{equation}\label{eq:log_odds_ratio}
    \text{Bias}_{\operatorname{PMI}}
        \approx \log \, \frac{\frac{f_{A,C}}{f_{A,nC}}}{\frac{f_{B,C}}{f_{B,nC}}}
        \approx \log \, \text{OR},
\end{equation}
where $\text{OR}$ is the odds ratio. Therefore, parametric confidence intervals and hypothesis testing can be conducted for $\text{Bias}_{\operatorname{PMI}}$ (details in Appendix \ref{sec:pmi_inference}). 

\section{Experiments} \label{sec:experiments}

To compare $\text{Bias}_{\operatorname{PMI}}$ with $\text{Bias}_{\text{WE}}$ we replicate three experiments that compare the gender biases measured in texts with the ones from other datasets:
\begin{enumerate} [leftmargin=*] 
\item \emph{Occupations-gender} \cite{caliskan2017semantics}: gender bias in text is compared to the percentage of women employed in a list of occupations in the U.S. Bureau of Labor Statistics in 2015.
\item \emph{Names-gender} \cite{caliskan2017semantics}: for a list of androgynous names, gender bias in text is compared to the percentage of people with each name who are women in the 1990 U.S. census.
\item \emph{Norms-gender} \cite{lewis2020gender}: textual gender bias is compared to the Glasgow Norms, a set of ratings for 5,500 English words which summarize the answers of participants who were asked to rate the gender association of each word \citep{scott2019glasgow}.
\end{enumerate}
Details about these datasets are in Appendix \ref{sec:datasets_details}.

We train GloVe, SGNS and PMI on two corpora: the 2014 English Wikipedia and English subtitles from OpenSubtitles \citep{lison2016opensubtitles}. We pre-process both corpora by converting all text to lowercase, removing non alpha-numeric symbols and applying sentence splitting, so that one sentence equates to one document. After pre-processing, the Wikipedia corpus is made up of 1.2 billion tokens and 53.9 million documents, whereas the OpenSubtitles corpus contains 2.4 billion tokens and 447.9 million documents. Refer to Appendix \ref{sec:corpora_details} for additional details about each corpus and to Appendix \ref{sec:model_details} for implementation details.
 
For each of the three settings, we assess the correlation between the dataset's female metric and the female bias as measured by PMI (equation \ref{eq:PMI_estimate}), and SGNS and GloVe (equation \ref{eq:BiasCaliskan}). Female bias refers to the bias metrics where $A$ and $B$ represent lists of female and male words, respectively.\footnote{
A=\textit{\{female, woman, girl, sister, she, her, hers, daughter\}} and
B=\textit{\{male, man, boy, brother, he, him, his, son\}} \citep{caliskan2017semantics,lewis2020gender}.
} Positive values imply that the target word is more associated with female terms than with male ones.

We measure correlation with Pearson's \emph{r}. We also compute a weighted Pearson's \emph{r}, which takes into account the standard error of each bias estimate and reduces the influence of noisy estimates on the correlation. Finally, for each word in each experiment we compute confidence intervals and p-values for the null hypothesis of absence of bias.\footnote{
In the case of $\text{Bias}_{\text{WE}}$, we apply bootstrap with 2,000 iterations and permutations with the all the possible combinations.
}

The aim of these experiments is not to find which method produces greater correlations in each task; it is rather to check whether $\text{Bias}_{\operatorname{PMI}}$ produces similar results to the widely used $\text{Bias}_{\text{WE}}$. If it does, it means our metric can extract trends from large corpora that correlate with gender stereotypes at least as well as embedding-based metrics can.

\section{Results} \label{sec:results} 

Table \ref{tab:correlations} shows Pearson's \emph{r} weighted and unweighted coefficients for each of the eighteen experiments (three association tests in two corpora with three bias measures each). The scatter plots associated with the Wikipedia's coefficients are available in Appendix \ref{sec:full_experiments}.

\input{correlations_table}

All in all, $\text{Bias}_{\operatorname{PMI}}$ and $\text{Bias}_{\operatorname{WE}}$ yield comparable results in these settings. There is no single method which consistently has the largest or lowest correlations. 

Weights tend to either increase the correlation considerably or to make it slightly weaker. This implies that in these experiments, noisy textual bias estimates usually agree less with the gender bias in the validation datasets. However, this does not mean that for each individual bias estimate the standard errors of each method are mutually interchangeable or equally useful (see section \ref{sec:inference}).

\begin{figure}[!ht]
    \centering
    \includegraphics[width=\linewidth]{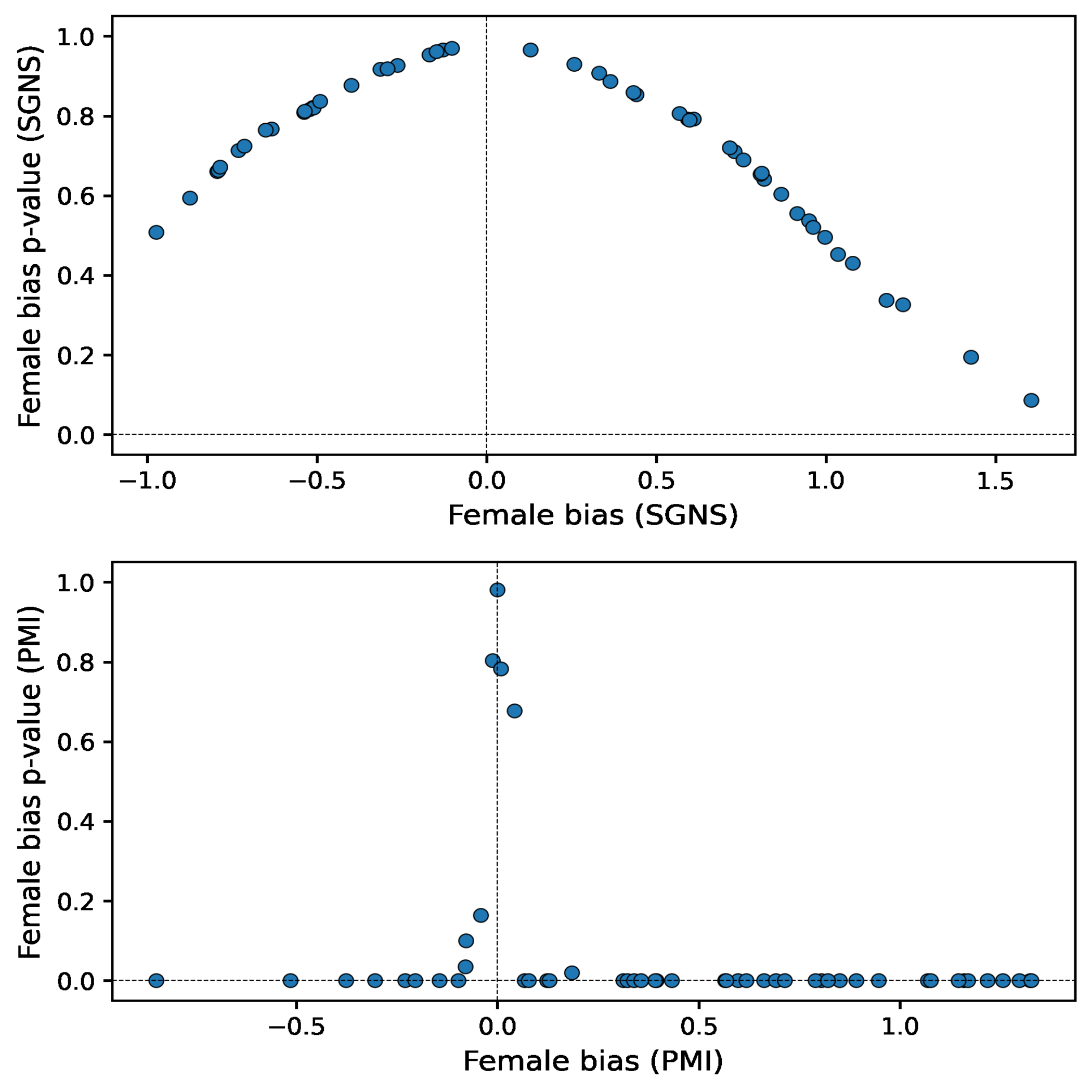}
    \caption{permutation p-values of $\text{Bias}_{\text{WE}}$ with SGNS vs. the value of $\text{Bias}_{\text{WE}}$ with SGNS (top panel), and $\log \text{OR}$ test p-values of $\text{Bias}_{\operatorname{PMI}}$ vs. the value of $\text{Bias}_{\operatorname{PMI}}$ (bottom panel) of androgynous names in Wikipedia.}
    \label{fig:pvalues}
\end{figure}

In Figure \ref{fig:pvalues} we compare the p-values of the permutation test of $\text{Bias}_{\text{WE}}$ with SGNS, with the p-values of the log odds ratio test of $\text{Bias}_{\operatorname{PMI}}$ for the \emph{Names-gender} test conducted in Wikipedia. A Benjamini-Hochberg correction was applied to the p-values obtained by both methods to account for multiple comparisons \citep{benjamini1995correction}. Appendix \ref{sec:all_pvalues} shows this example is consistent with the rest of the experiments.

In this example, only the word with the highest $\text{Bias}_{\text{WE}}$ is significantly different from zero at a 0.10 significance level. In contrast, most words have a $\text{Bias}_{\operatorname{PMI}}$ significantly different from zero, with the exception of some points with bias values close to zero. This is because the procedures that compute p-values for each type of metric capture essentially different types of variability (see section \ref{sec:inference}).

\section{Discussion} \label{sec:discussion}

\subsection{Interpretability}

Although there are studies on how word vector spaces are formed \citep{levy2014neural,levy2015improving,ethayarajh2019understanding} and on the biases they encode \citep{bolukbasi2016man,zhao2017men,gonen2019lipstick}, there is no transparent interpretation of the embedding-based bias metrics in terms of co-occurrences of words in the texts.

In contrast, $\text{Bias}_{\operatorname{PMI}}$ can be expressed intrinsically in terms of conditional probabilities (equation \ref{eq:PMI_estimate}). The bias is interpreted as the logarithm of how much more likely it is to find words in $C$ in the context of words in $A$ than in the context of words in $B$. For example, in the Wikipedia corpus the female $\text{Bias}_{\operatorname{PMI}}$ of word \emph{nurse} is 1.3172, thus,
\begin{equation*}
    \frac{P(nurse|A)}{P(nurse|B)} = \exp \, 1.3172 = 3.7330.
\end{equation*}
This means that it is 273.30\% more likely to find the word \emph{nurse} in the context of female words ($A$) than in the context of male words ($B$).

To the lack of interpretability of $\text{Bias}_{\text{WE}}$ contributes the fact that SGNS and GloVe can capture word associations of second order or higher \citep{altszyler2018corpus,schlechtweg2019second}, whereas $\operatorname{PMI}$ is strictly a first-order association metric. When embeddings are used to measure biases, it is not possible to tell whether the results are due to widespread first-order co-occurrences or are derived from obscure higher-order co-occurrences \citep{brunet2019understanding,rekabsaz2021measuring}.

For instance, in OpenSubtitles, the $\text{Bias}_{\text{PMI}}$ of the word \emph{evil} equals $-0.25$, indicating a higher likelihood of appearing in the context of male context words ($B$) compared to female ones ($A$). Conversely, $\text{Bias}_{\text{SGNS}}=0.23$. Even if this stands for female bias, it is difficult to understand the exact source of this result since it is influenced by second and higher-order co-occurrences. Moreover, in recent research we demonstrated that $\text{Bias}_{\text{WE}}$ can also yield misleading results by inadvertently capturing disparities in the frequencies of context words \citep{valentini2022undesirable}.






Nevertheless, bias metrics that capture second-order associations have the advantage of managing data sparsity. Since word embeddings can capture synonymy, when data is sparse it might not be necessary to include all related words to the concepts of interest in order to measure meaningful biases. In the case of our first-order metric, this problem must be addressed by increasing word lists with synonyms and forms of the words of interest.

To illustrate this, let's consider the case of the words \emph{nourish} and \emph{nurture}, which have different frequencies in the Wikipedia corpus ($700$ and $3,000$, respectively). With $\text{Bias}_{\text{PMI}}$, we obtain a bias of $0.33$ for \emph{nurture} (p-value $< 10^{-4}$). However, if we had used its less frequent synonym \emph{nourish} instead, the $\text{Bias}_{\text{PMI}}$ would have been $-0.10$ and not statistically significant (p-value $\approx 0.66$). Here we would not have been able to determine whether there is actually no bias or if there is insufficient data. This shows that it is generally advisable
to include all pertinent synonyms and variations of the term whose bias we are trying to measure.




\subsection{Statistical inference} \label{sec:inference}

The p-values, standard errors and confidence intervals of the $\log \text{OR}$ approximation are fundamentally different from the ones estimated for $\text{Bias}_{\text{WE}}$ through permutations and bootstrap. The uncertainty quantified for $\text{Bias}_{\operatorname{PMI}}$ captures the variability of the underlying data generating process i.e. the one induced by the randomness of co-occurrence counts as random quantities. In contrast, the estimates for $\text{Bias}_{\operatorname{WE}}$ only consider the variability across the sets of context words. This means that multiple words \emph{must} be chosen so that inference can be conducted. In fact, whenever $A$ and $B$ are single-word lists, there is no way of estimating uncertainty for $\text{Bias}_{\operatorname{WE}}$ with these methods, whereas it is perfectly feasible for $\text{Bias}_{\operatorname{PMI}}$.

As far as we know, we are the first to provide a simple and efficient way of evaluating the statistical significance of bias. This is especially important in Computational Social Science, for which it is useful to have not only a reliable metric to quantify stereotypes but also a reliable tool to measure uncertainty i.e. to know up to what degree the measured values might have been due to statistical fluctuation. Meaningful statistical tests and confidence intervals that capture the variability that really matters are therefore essential.

\section{Conclusion}

We presented a PMI-based metric to quantify biases in texts, which (a) allows for simple and computationally inexpensive statistical inference, (b) has a simple interpretation in terms of word co-occurrences, and (c) is explicit and transparent in the associations that it is quantifying, since it captures exclusively first-order co-occurrences. Our method produces similar results to the GloVe-based and SGNS-based metrics in experiments which compare gender biases measured in large corpora to the gender gaps of independent empirical data. 

\section*{Limitations}

We replicate three well-known experiments in the gender bias literature, where bias is measured according to a binary female vs. male view. This choice ignores other views of gender but eases the presentation of the frameworks.

We only use two corpora and three datasets which by no means capture the biases of all the people speaking or writing in the English language. Moreover, we don't experiment with different corpus sizes, a more diversified set of corpora or more bias types. We hope to explore this in future work.

The hyperparameters of the models have not been varied, using their default values. This replicates the standard experimental setting used in the literature. Since there are no ground truths when measuring biases (that is, there are no annotations with the amount of bias of words in large corpora), hyperparameters are usually set to their default values.

\bibliography{acl2023}
\bibliographystyle{acl_natbib}

\appendix

\section{Contingency table of co-occurrences}\label{sec:contingency_table}

$\text{Bias}_{\operatorname{PMI}}$ is computed with the co-occurrences between the groups of words $A$, $B$ and $C$. These can be represented with the following contingency table:

\begin{table}[H]
\centering
\begin{tabular}{|c|c|c|c|} 
 \hline
   & $C$ &  $not \; C$ & Total \\ \hline
 $A$  & $f_{A,C}$ & $f_{A,nC}$ & $f_{A,C}$+$f_{A,nC}$ \\ \hline
 $B$ & $f_{B,C}$ & $f_{B,nC}$ & $f_{B,C}$+$f_{B,nC}$ \\ \hline
\end{tabular}
\caption{Contingency table of words co-occurrences}
\label{tab:contingency}
\end{table}
This contains, for example, how many times words in $A$ appear in the context of words in $C$ ($f_{A,C}$) and how many times they do not ($f_{A,nC}$).

\section{Statistical inference for $\text{Bias}_{\operatorname{PMI}}$}\label{sec:pmi_inference}

The distribution of the log odds ratio (equation \ref{eq:log_odds_ratio}) converges to normality \citep{agresti2003categorical}. Its 95\% confidence interval is given by
\begin{equation*}
    CI_{95\%}\big( \text{Bias}_{\operatorname{PMI}} \big) = 
        \text{Bias}_{\operatorname{PMI}} \pm 1.96 \, SE
\end{equation*}
with
\begin{eqnarray*}\label{eq:SE}
    SE &=& \sqrt{\frac{1}{f_{A,C}}+\frac{1}{f_{B,C}}+\frac{1}{f_{A,nC}}+\frac{1}{f_{B,nC}}}\\
    &\approx&  \sqrt{\frac{1}{f_{A,C}}+\frac{1}{f_{B,C}}}
\end{eqnarray*}
This last approximation considers condition \ref{eq:approx}.

We can test the null hypothesis that the log odds ratio is 0 (absence of bias) with a standard Z-test, whereby the two-sided p-value is computed with $2P(\text{Z} < -|\text{Bias}_{\operatorname{PMI}}|/SE)$, where Z is a standard normal random variable.

\section{Datasets}\label{sec:datasets_details}

For the \emph{occupations-gender} and \emph{names-gender} experiments, the female proportions for names and occupations in the U.S. were extracted from the datasets provided by Will Lowe's \texttt{cbn} R library \footnote{\url{https://conjugateprior.github.io/cbn/}}, which contains tools for replicating \citet{caliskan2017semantics}. We used the 50 names and 44 occupations available in this source. 

The original Glasgow Norms comprise 5,553 English words. Individuals from the University of Glasgow were asked to measure the degree to which each word is associated with male or female behavior on a scale from 1 (very feminine) to 7 (very masculine). Following \citet{lewis2020gender}, we average the norms of homonyms and compute $8 - rating$ to flip the scale so that it represents \emph{femaleness} according to human judgement. 4,668 words from the original list overlapped with OpenSubtitle's vocabulary, and 4,642 words overlapped with the Wikipedia vocabulary.

\section{Corpora}\label{sec:corpora_details}

The Wikipedia corpus was built from the August 2014 dump, licensed under CC BY-SA 3.0\footnote{\url{https://archive.org/download/enwiki-20141208}}. We removed articles with less than 50 tokens.

The OpenSubtitles corpus \citep{lison2016opensubtitles} includes English subtitles from movies and TV shows and was built with the aid of the \texttt{subs2vec} Python package with MIT License \citep{vanparidon2021subs}.


\section{Model training}\label{sec:model_details}

We ignore words with less than 100 occurrences, resulting in a vocabulary of 172,748 words for Wikipedia and 128,974 words for OpenSubtitles. 

We use a window size of 10 in all models and apply "dirty" subsampling i.e. out-of-vocabulary tokens are removed before the corpus is processed into word-context pairs \citep{levy2015improving}.

Word embeddings with 300 dimensions are trained with SGNS and GloVe. For SGNS we use the word2vec implementation of Gensim 4.1.2 licensed under GNU LGPLv2.1 \citep{rehurek2010gensim} with default hyperparameters. GloVe is trained with the original implementation \citep{pennington2014glove} with version 1.2 (Apache License, Version 2.0) with 100 iterations. This version uses by default additive word representations, in which each word embedding is the sum of its corresponding context and word vectors. 

For PMI, we count co-occurrences with the GloVe module \citep{pennington2014glove} with version 1.2 and set the smoothing parameter $\epsilon$ to 0.5. 

We ran all experiments on a desktop machine with 4 cores Intel Core i5-4460 CPU and 32 GB RAM. Training times were around 1 hour per epoch with SGNS and 5 minutes per iteration with GloVe. Co-occurrence counts used for PMI were obtained in around 20 minutes with GloVe. 

\section{Results}\label{sec:full_results}

\subsection{Experiments}\label{sec:full_experiments}

\begin{figure*}[!ht]
  \begin{center}
    \includegraphics[width=\linewidth]{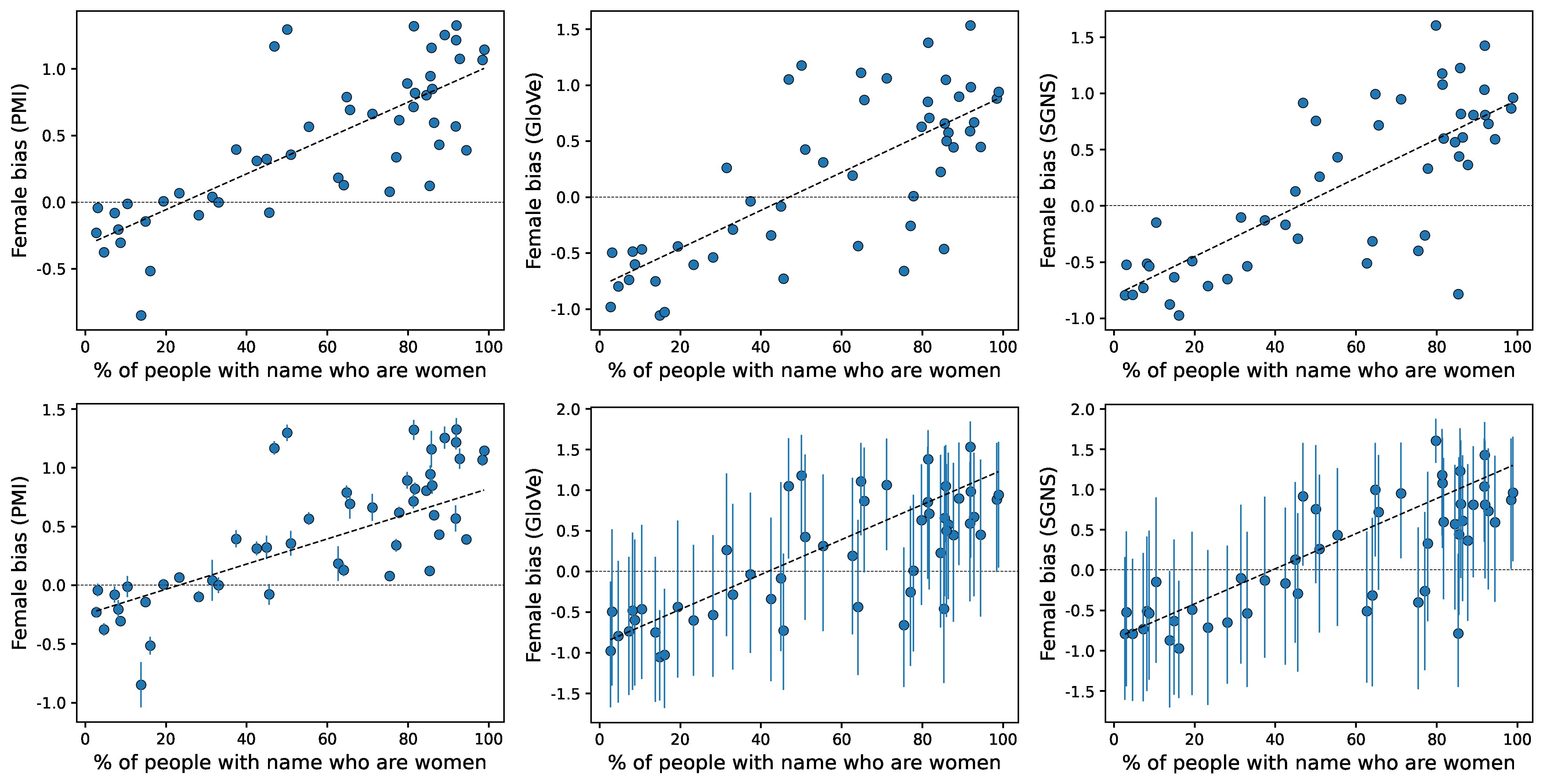}
  \end{center}
  \caption{\emph{Names-gender} experiments in Wikipedia. Horizontal axes represent the percentage of people with each name who are women as measured in the 1990 U.S. census.}
  \label{fig:scatter_names}
\end{figure*}

\begin{figure*}[!ht]
  \begin{center}
    \includegraphics[width=\linewidth]{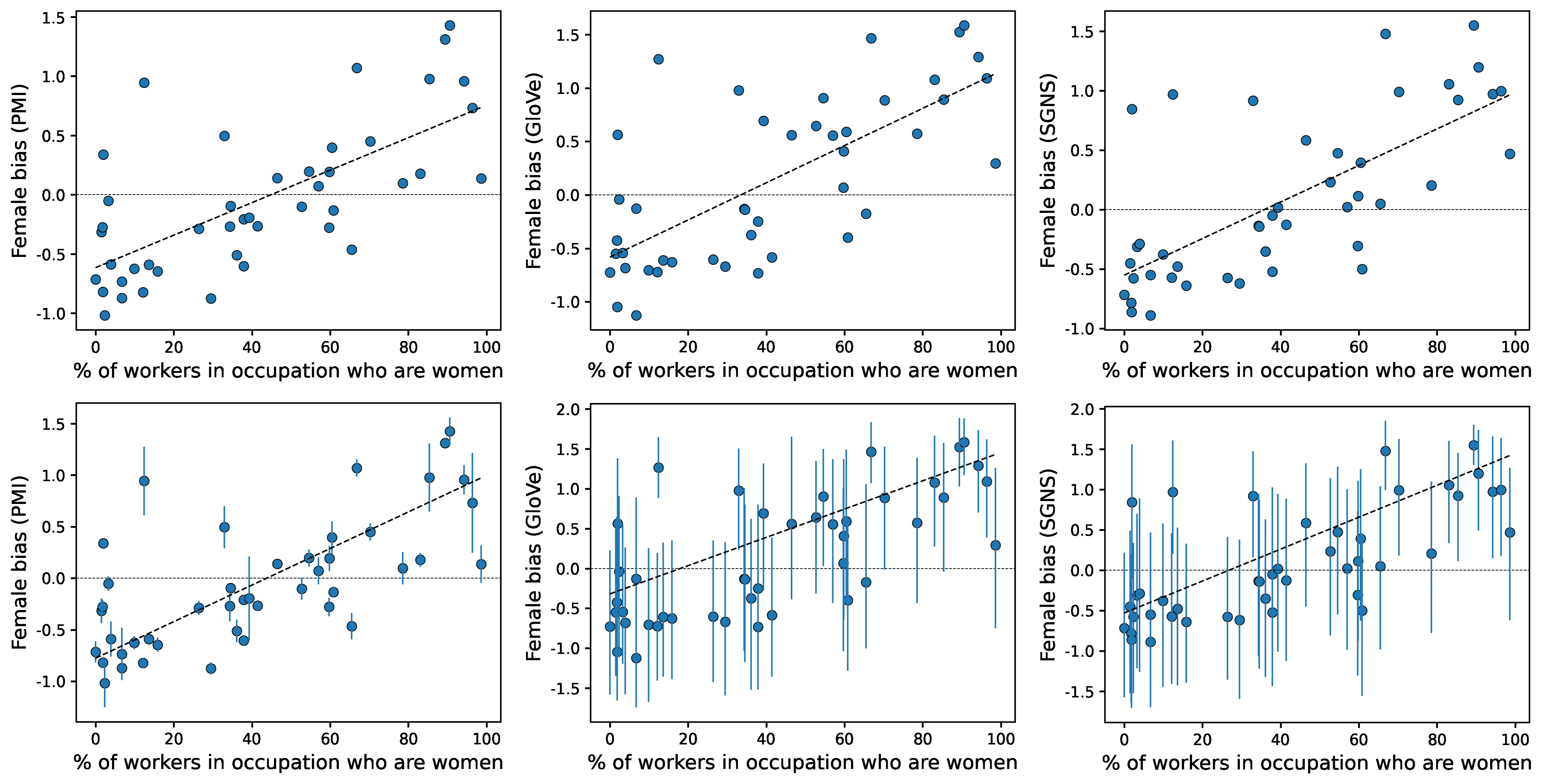}
  \end{center}
  \caption{\emph{Occupations-gender} experiments in Wikipedia. Horizontal axes represent the percentage of women employed in each occupation in 2015 according to the U.S. Bureau of Labor Statistics.}
  \label{fig:scatter_occupations}
\end{figure*}

\begin{figure*}[!ht]
  \begin{center}
    \includegraphics[width=\linewidth]{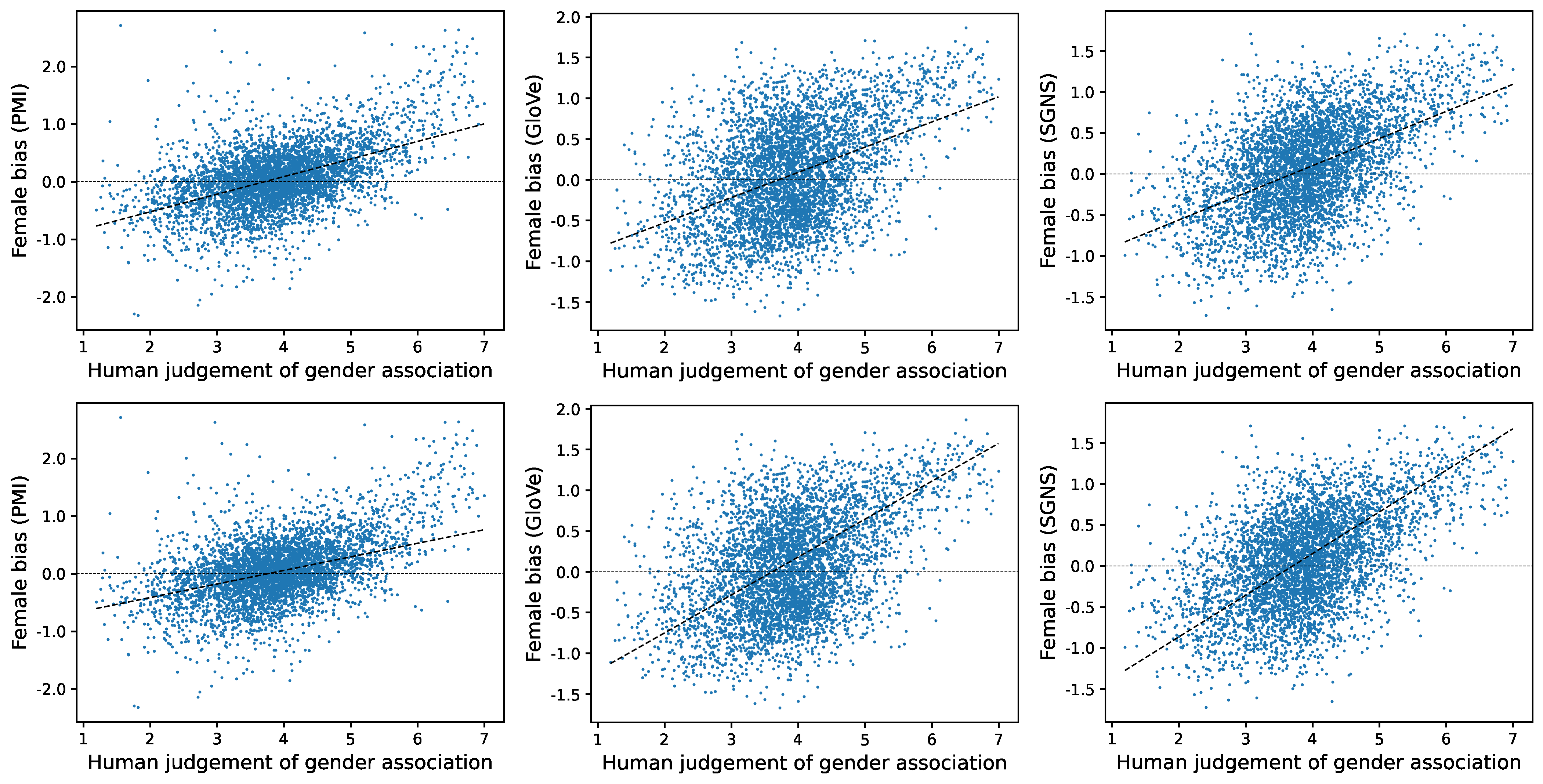}
  \end{center}
  \caption{\emph{Norms-gender} experiments in Wikipedia. Horizontal axes represent the Glasgow Norm of each word. Confidence intervals are not displayed in the second row to avoid overplotting.}
  \label{fig:scatter_norms}
\end{figure*}

In Figures \ref{fig:scatter_names}, \ref{fig:scatter_occupations} and \ref{fig:scatter_norms} we display the scatter plots of the three experiments described in section \ref{sec:experiments} for the Wikipedia corpus. The findings for OpenSubtitles are qualitatively the same and we exclude the plots for simplicity. 

The vertical axes represent the  female vs. masculine bias measures based on PMI (left panels), GloVe (middle panels), and SGNS (right panels). 
Dashed lines represent linear regressions. In the second row, the bias standard error was taken into account as weights in the regression, and error bars are confidence intervals.

All unweighted and weighted correlation coefficients in Table \ref{tab:correlations} are significantly different from zero at the 0.0001 level.

\subsection{p-values}\label{sec:all_pvalues}

\begin{figure*}[!ht]
  \begin{center}
    \includegraphics[width=\linewidth]{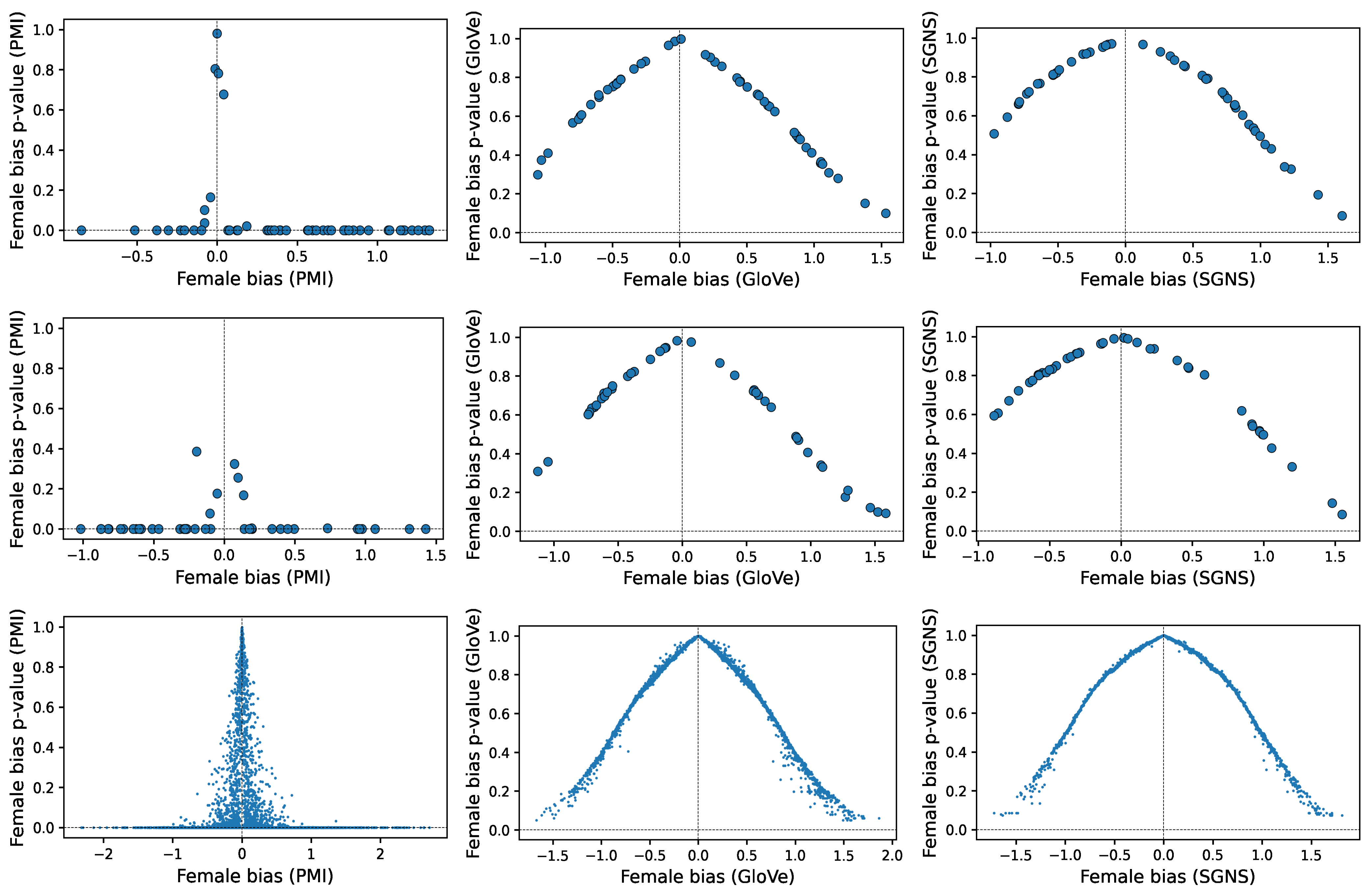}
  \end{center}
  \caption{Female vs. male bias p-values of the \emph{names-gender} (row 1), \emph{occupations-gender} (row 2) and \emph{norms-gender} (row 3) experiments in Wikipedia.
  }
  \label{fig:scatter_pvalues}
\end{figure*}

Figure \ref{fig:scatter_pvalues} shows the corrected p-values for the gender bias of each word in the vertical axes vs. the value of the bias in the horizontal axes. p-values for SGNS and GloVe result from permutations tests whereas PMI uses the log odds ratio test. All p-values have been corrected with Benjamini-Hochberg separately for each setting. The plots for OpenSubtitles are very similar and are excluded for simplicity. 

\end{document}

%% file: correlations_table.tex
\begin{table*}[t]
\centering

\begin{tabular}{llcrrr}
\toprule
\textbf{Corpus} & \textbf{Experiment} & \textbf{Correlation} &  \textbf{PMI} & \textbf{GloVe} & \textbf{SGNS} \\

\midrule
\multirow[c]{6}{*}{OpenSubtitles} & \multirow[c]{2}{*}{Glasgow-Gender} & $r$ & 0.58 & 0.49 & 0.55 \\
 &  & Weighted $r$ & 0.58 & 0.69 & 0.72 \\
\cline{2-6}
 & \multirow[c]{2}{*}{Names-Gender} & $r$ & 0.80 & 0.74 & 0.81 \\
 &  & Weighted $r$ & 0.84 & 0.82 & 0.77 \\
\cline{2-6}
 & \multirow[c]{2}{*}{Occupations-Gender} & $r$ & 0.66 & 0.67 & 0.79 \\
 &  & Weighted $r$ & 0.81 & 0.83 & 0.89 \\
\cline{1-6} \cline{2-6}
\multirow[c]{6}{*}{Wikipedia} & \multirow[c]{2}{*}{Glasgow-Gender} & $r$ & 0.50 & 0.44 & 0.50 \\
 &  & Weighted $r$ & 0.44 & 0.59 & 0.66 \\
\cline{2-6}
 & \multirow[c]{2}{*}{Names-Gender} & $r$ & 0.78 & 0.74 & 0.77 \\
 &  & Weighted $r$ & 0.75 & 0.79 & 0.76 \\
\cline{2-6}
 & \multirow[c]{2}{*}{Occupations-Gender} & $r$ & 0.69 & 0.70 & 0.70 \\
 &  & Weighted $r$ & 0.79 & 0.67 & 0.78 \\
\cline{1-6} \cline{2-6}
\bottomrule

\end{tabular}

\caption{Pearson's \emph{r} coefficients of each experiment. Weighted \emph{r} accounts for the variability of each bias estimate.}
\label{tab:correlations}
\end{table*}